\pdfoutput=1

\documentclass[11pt]{article}

\usepackage{ACL2023}

\usepackage{times}
\usepackage{latexsym}

\usepackage{amsmath} 
\usepackage{booktabs} 

\usepackage[T1]{fontenc}

\usepackage[utf8]{inputenc}

\usepackage{microtype}

\usepackage{inconsolata}

\usepackage{fontawesome}    
\usepackage{graphicx}
\usepackage{bbm}

\title{Rewarding Chatbots for Real-World Engagement with Millions of Users}

\author{
Robert Irvine
\quad
Douglas Boubert $^{\ast,\dagger}$
 \quad Vyas Raina $^{\ast,\ddagger}$
\quad  Adian Liusie $^{\ast,\ddagger}$ \\
\AND
Ziyi Zhu
\quad
Vineet Mudupalli
\quad 
Aliaksei Korshuk
\quad
Zongyi Liu 
\quad
Fritz Cremer
\AND
Valentin Assassi$^{\dagger}$
\quad
Christie-Carol Beauchamp
\quad
Xiaoding Lu
\AND
Thomas Rialan
\quad 
William Beauchamp
\\\\
Chai Research
}

\begin{document}

\maketitle

\begin{NoHyper}
\def\thefootnote{*}\footnotetext{Equal contribution}
\def\thefootnote{$\dagger$}\footnotetext{Seamless Capital}\def\thefootnote{\arabic{footnote}}
\def\thefootnote{$\ddagger$}\footnotetext{Machine Intelligence Laboratory, University of Cambridge}\def\thefootnote{\arabic{footnote}}
\end{NoHyper}

\begin{abstract}
The emergence of pretrained large language models has led to the deployment of a range of social chatbots for chitchat. Although these chatbots demonstrate language ability and fluency, they are not guaranteed to be engaging and can struggle to retain users. This work investigates the development of social chatbots that prioritize user engagement to enhance retention, specifically examining the use of human feedback to efficiently develop highly engaging chatbots. The proposed approach uses automatic pseudo-labels collected from user interactions to train a reward model that can be used to reject low-scoring sample responses generated by the chatbot model at inference time. Intuitive evaluation metrics, such as mean conversation length (MCL), are introduced as proxies to measure the level of engagement of deployed chatbots. A/B testing on groups of 10,000 new daily chatbot users on the Chai Research platform shows that this approach increases the MCL by up to 70\%, which translates to a more than 30\% increase in user retention for a GPT-J 6B model. Future work aims to use the reward model to realise a data fly-wheel, where the latest user conversations can be used to alternately fine-tune the language model and the reward model~\footnote{Models and Code open sourced at: \url{https://huggingface.co/ChaiML}}.
\end{abstract}

\section{Introduction}
The recent surge in pretrained large language models (PrLMs; ~\citealp{zhu2022simple}) has transformed the natural language processing field by enabling systems to perform a variety of language tasks with human-like proficiency. One such task is dialogue generation, where users can interact with social chatbots~\cite{10.1145/3166054.3166058} in a conversational chitchat setting. However, while PrLM-generated responses are often coherent and on-topic, they may not always be engaging, leading to shorter conversations and lower user retention. This paper focuses on explicitly developing social chatbots that prioritize user engagement to enhance retention.


Recent work has shown that human feedback is a very promising and effective method to align systems with human intent. \citet{ouyang2022training} show that by having annotators explicitly rank preferred responses, reinforcement learning from human feedback \citep[RLHF;][]{christiano2017deep, stiennon2020learning} can be used to improve language models such that responses are more \textit{helpful}, \textit{honest}, and \textit{harmless}. These solutions, however, require manual annotations where fluent contractors have to manually rank tens of thousands of responses per system. This process can be very time consuming and expensive, which can limit progress in developing and evaluating chatbots. Additionally, explicitly ranking responses based on their level of engagement can be challenging for human annotators. Therefore, this paper proposes a more efficient approach for collecting feedback by using automatic pseudo-labels that serve as good proxies for user engagement, collected during user interactions with the chatbot. These pseudo-labels are used to train a reward model, which can then be used to reject low-scoring sample responses generated by the chatbot model at inference time~\cite{DBLP:journals/corr/abs-1912-02164}.

Further, this work proposes intuitive evaluation metrics that directly measure how engaging deployed chat-bots are, and show that our proposed method significantly improves the level of engagement of a GPT-J 6B \citep{gpt-j} based chatbot. Through A/B tests on groups of 10,000 new daily chatbot users, we show that our method increases the retention of a GPT-J 6B model by more than 30\%, highlighting the effectiveness of using human feedback for developing engaging chatbots. We run all of our evaluation on the Chai Research\footnote{\url{https://www.chai-research.com/}} platform, an online app with \textit{millions} of daily users, where users can chat with chatbots designed to act as friends, mentors or fictional characters. We publicly release all of the anonymised user conversations and pseudo labels used in this paper, with the hope that this resource can stimulate further interest in developing highly engaging and entertaining chatbots.

\begin{figure}[h]
  \centering
  {%
    \setlength{\fboxsep}{0pt}%
    \setlength{\fboxrule}{1pt}%
    \fbox{\includegraphics[width=0.7\columnwidth]{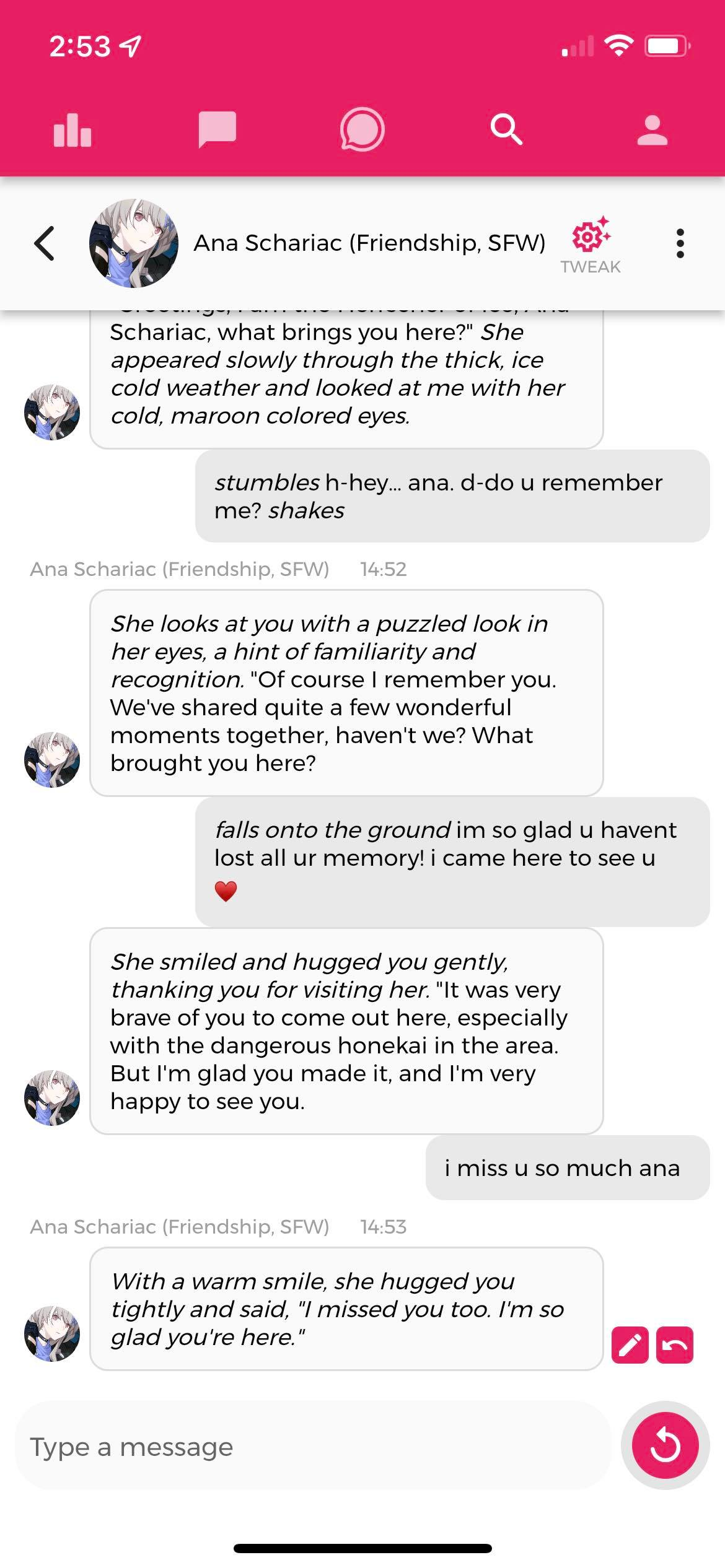}}%
  }%
  \caption{Screenshot of one of the authors interacting with a highly engaging chatbot on the Chai platform.}
  \label{fig:screenshot}
\end{figure}
\section{Related Work}

\textbf{Chatbot Designs:}
Chatbots and dialogue systems are designed for many applications, ranging from virtual assistants responding to goal-oriented user queries, to social chatbots designed for casual chitchat with a human user~\cite{10.1145/3166054.3166058}. This work focuses on chatbots for chitchat, where the objective is to provide user entertainment and engagement. Early social chatbots that used rule-based methods~\cite{10.1145/365153.365168} were followed by retrieval-based models, that more recently have been combined with various generative models~\cite{nlp4convai-2021-natural}. With the emergence of pre-trained large language models~\cite{zhu2022simple}, social chatbots have recently been dominated by transformer-based \cite{NIPS2017_3f5ee243} designs~\cite{10.1145/3373017.3373028}. Typically, transformer-based chatbots are fine-tuned on conversational data in a specific domain, e.g. GPT-2 finetuned on conversational reddit data \citet{zhao-etal-2022-unids}. Over the last few years, larger and better transformer-based models have been trained on conversational data for the development of more sophisticated social chatbots~\cite{ DBLP:journals/corr/abs-2001-09977, roller-etal-2021-recipes, DBLP:journals/corr/abs-2006-16779, choudhary-kawahara-2022-grounding, yan2022deep}.

\noindent\textbf{Learning using Human Feedback:}
Instead of purely fine-tuning large language models on conversational data, incorporating human feedback in chatbot development has been shown to be an effective method to obtain more human-aligned responses~\cite{DBLP:journals/corr/abs-1811-07871, DBLP:journals/corr/abs-2112-00861, DBLP:journals/corr/abs-2001-09768}. Popularized by the training of InstructGPT~\cite{ouyang2022training}, reinforcement learning from human feedback \citep[RLHF;][]{christiano2017deep, stiennon2020learning} trains a reward model on user ranks of system responses, followed by reinforcement learning as an agent interacting with the environment. As an alternative to reinforcement learning where model parameters are updated by PPO, the reward model can also be used at inference time by, for instance, rejecting the \textit{worst} responses generated by the model~\citep{DBLP:journals/corr/abs-1912-02164}. Human feedback has also been used in several other ways including constrained optimization~\cite{DBLP:journals/corr/AchiamHTA17}, control codes~\cite{DBLP:journals/corr/abs-1909-05858} and expert iteration~\cite{DBLP:journals/corr/AnthonyTB17, DBLP:journals/corr/abs-1712-01815}.

\noindent\textbf{Chatbot Evaluation:}
The traditional method of assessing chatbots is through direct human assessment. This can be done through rating individual responses \citep{DBLP:journals/corr/RadziwillB17, DBLP:journals/corr/abs-1805-08455, sordoni-etal-2015-neural}, scoring aggregated qualities across the multi-turn exchange, or through ranking overall conversational experiences \citep{deriu-etal-2020-spot, DBLP:journals/corr/abs-1909-03087, 10.1145/175208.175217}. Alternatively, text overlap based metrics such as BLEU, METEOR and ROUGE~\cite{sordoni-etal-2015-neural, 10.1007/978-3-319-77113-7_25} and F-score metrics~\cite{DBLP:journals/corr/abs-2010-07079, DBLP:journals/corr/abs-1908-10422} have also been proposed to assess chatbots. Other assessment methods \citep{yang-etal-2022-chatmatch} include sentence perplexity~\cite{DHYANI2021817, 10.1007/978-3-319-59569-6_32, higashinaka-etal-2014-towards}, entities per exchange~\cite{finch-choi-2020-towards}, number of questions raised, specificity~\cite{li-etal-2016-diversity}, turns per conversation~\cite{DBLP:journals/corr/abs-1801-01957}, inconsistency detection, and relevance to history. In this work, we argue that the inherent goal of deployed social chatbots for chitchat is to be engaging and entertaining. Text overlap scores do not consider that agents can give many high quality responses outside the reference set, while explicit human feedback can be expensive, time consuming, and limit the scope of analysis. In this paper we therefore consider alternative evaluation metrics such as user retention and average conversational length, which are better suited as metrics for the properties we are interested in.

\section{Evaluating Chit Chat Bots} \label{sec:conversability}
Chatbots have been deployed for a range of applications, including in task oriented dialogue systems and as virtual assistants. This work, however, is concerned with building entertaining social chatbots, where the system keeps the user engaged in an engrossing conversation, such that the user enjoys the experience and is thus encouraged to continue the conversation. We believe that the critical property to optimise for social chatbots is their level of \textit{engagement} and \textit{entertainment}. Due to the open-domain nature of the task, text overlap scores are clearly an ineffective evaluation metric that may falsely penalise creative and interesting responses. Though traditional explicit human evaluation remains plausible for assessing systems, doing so is subjective, expensive and hard to scale. 

We instead make the simple observation that within user conversations, there already exist clear signals that demonstrate the quality of how engaging the deployed social bot is. Our metrics are founded on the simple assumption that if users find the chatbot's responses captivating, they are likely to converse with the system for longer, and are also more likely to return to the platform to converse in future (i.e. have a higher user retention). As such, we propose several simple and effective methods for evaluating deployed chatbots, aligned to the overall goal of user engagement, directly and objectively.

\subsection{Engagement Evaluation Metrics}
\label{sec:metrics}

\textbf{Mean Conversation Length}
The mean conversation length measures the average number of user queries per conversation session. Let each conversation session $\mathcal{C}$ be an ordered set of user and system response pairs (conversation \textit{turns}),
\begin{align}
    \mathcal{C} = \{(u_1, r_1),  (u_2, r_2), \dots, (u_N, r_N)\}.
\end{align}
Over a set of independently drawn multiple sessions $\mathcal{D} = \{\mathcal{C}_1, \mathcal{C}_2, \dots\}$, the \textit{mean conversion length} can be calculated as:
\begin{equation}\label{eqn:mcl}
    \operatorname{MCL}(\mathcal{D}) = \frac{1}{|\mathcal{D}|} \sum_{\mathcal{C} \in \mathcal{D}} |\mathcal{C}|.
\end{equation}
Conversation length statistics for the Chai Research platform are given in Appendix \ref{sec:conversation_length}.\newline

\noindent\textbf{Retry Rate}
A common functionality present in chatbot interfaces is a button that allows the user to request the chatbot to regenerate an alternate response to their prompt. Let $\mathcal R$ be the entire set of all responses $r\in\mathcal C$, for all conversations $\mathcal C\in\mathcal D$. Further, let the function $G:r \rightarrow \{0, 1\}$ denote whether response $r$ has been regenerated at least once. Then \textit{retry rate} is the fraction of system responses that the user requested to regenerate at least once,
\begin{equation}\label{eqn:retry}
    \operatorname{RetryRate}(\mathcal{D}) = \frac{1}{|\mathcal R|}\sum_{r\in\mathcal R}G(r).
\end{equation}

\noindent\textbf{User Star Rating}
Typical chatbot platforms periodically~\footnote{The Chai platform requests user ratings for 5\% of responses} request users to provide feedback by rating a specific response $r$ on a rating scale, for example, from one (worst) to four (best) stars. Let $\mathcal R_*$ be the set of all responses $r\in\mathcal C$ for all conversations $C\in\mathcal D$ that received a user star rating and let the function $H:r \rightarrow \{0, 1\}$ denote whether response $r\in\mathcal R_*$ received at least $S$ stars. Then the $S$-Star Rate is simply the fraction of star ratings by users who responded to the survey that is $S$ stars or more.
\begin{equation}\label{eqn:four}
    \operatorname{StarRating}_{S}(\mathcal{D}) = \frac{1}{|\mathcal R_*|}\sum_{r\in\mathcal R_*}H(r).
\end{equation}

\noindent\textbf{Retention Rate}
Retention of users on the chatbot platform is the overall metric that commercial entities will often aim to optimise. Day $X$ retention can be defined as the fraction of users that engage with the chatbot on the $X$th day after their first conversation. It is typical to optimise the day thirty (D30) retention rates of users (on the Chai platform in this work). However, to obtain a measure of retention, the chatbot has to be deployed for at least $X$ days and that the initial group of users must interact with the same chatbot for all $X$ days\footnote{Users in a retention experiment can still interact with different \textit{characters} on the Chai Research platform. We stress the distinction between a chatbot and a character (a chatbot prompted to respond as a specific character in a specific context).}. Hence, retention rate is a challenging and expensive evaluation metric to use, despite being a desired overall measure of chatbot engagement.

\section{Method}
The overall approach we take is a three stage pipeline, inspired by InstructGPT~\cite{ouyang2022training}. The first stage is finetuning, where a PrLM is finetuned on specific conversational and literary data that is better suited to the downstream task. The second stage is reward model training, where we learn a reward model that can determine whether a given response is engaging and enjoyable for the user. The final stage then leverages the reward model at inference time so that the chatbot can produce responses that are considered captivating. 

\subsection{Model Fine-tuning}
PrLMs are trained on a wide variety of text sources. Despite strong zero-shot performance on specific tasks and text domains, finetuning these models has become the de facto standard to achieve greater domain-specific performance gains~\cite{DBLP:journals/corr/abs-2003-08271}. Therefore, this work also uses a standard next word prediction training scheme to finetune decoder-based pretrained large language models on specifically \textit{entertaining} text domains, e.g. literature.

\noindent\subsection{Reward modelling}
The aim is to incorporate human feedback in the design of more engaging chatbots. We propose to use a reward model $R$ that learns how engaging a response $r_i$ is. Let $y_i \in \{0, 1\}$ denote whether response $r_i$ was engaging given the back history of user prompts $u_{1:i}$ and system responses $r_{1:i-1}$. The reward model is trained to learn 
\begin{equation}
    R(r_i) = P(y_i| r_i, u_{1:i}, r_{1:i-1}).
\end{equation}

To learn a reward model using supervised training, one requires labeled data, where each response is marked with a measure of how engaging it is. Manual annotations can be expensive and laborious to collect, which can limit the size of a training dataset. Therefore it would be preferable to use proxy labels that are naturally present in user conversations, without requiring explicit human annotation. We therefore propose pseudo labels that can be conveniently extracted from user interactions, where the labels directly mark whether a given response is engaging or not. Section \ref{sec:metrics} considers some simple chatbot evaluation metrics to measure engagement and so, when designing pseudo-labels for training a reward model, we consider three different pseudo-labelling strategies, aligned to these engagement metrics.\\

\noindent\textbf{Conversational length:}
If the user finds the chatbot's response interesting, then it is reasonable to expect that the user will continue conversing. Therefore a very natural idea is to assume that the last few system responses were not engaging to the user, and that all previous ones were. Hence, for each conversation $\mathcal C$, we can label each response $r$ as follows,
\begin{equation}
    \begin{cases}
      y_i = 1, & \text{for} \quad i \in \{1, \dots, {N-K +1}\} \\
      y_i = 0, & \text{for} \quad i \in \{{N-K + 1}, \dots, N\},
    \end{cases}
\end{equation}
where $N$ is the conversation length. Here $K$ is a hyper-parameter that defines how many user messages must be sent subsequent to a chatbot response for that response to be classed as engaging. The special case of $K=1$ corresponds to classifying a message as engaging if the user replies at all and thus continues the conversation. Use of these pseudo-labels is naturally expected to directly optimise the mean conversation length, MCL (Equation \ref{eqn:mcl}) engagement evaluation metric.\\

\noindent\textbf{Retry ability:}
The Chai platform gives users the option to regenerate any responses they are not satisfied with. It can therefore be assumed that any regenerated responses are considered neither satisfactory nor engaging, while otherwise the responses are reasonable. As in Equation \ref{eqn:retry}, $G(r_i) \in \{0, 1\}$ denotes whether the user requested to regenerate response $r_i$ given the back history. Then we can define pseudo-labels for each response, $r$ in a conversation to align directly to the retry rate engagement evaluation metric,
\begin{equation}
    \begin{cases}
      y_i = 1, & \text{if} \quad G(r_i) = 0, \\
      y_i = 0, & \qquad \text{otherwise}.
    \end{cases}
\end{equation} \\

\noindent \textbf{Manual Labels:}
Seeking to optimise the four star engagement evaluation metric defined in Equation \ref{eqn:four}, we exploit explicit user ratings to construct pseudo-labels,
\begin{equation}
    \begin{cases}
      y_i = 1, & \text{if} \quad H(r_i) = 1, \\
      y_i = 0, & \qquad \text{otherwise}.
    \end{cases}
\end{equation}

\subsection{Best-of-$N$ Sample Rejection}
The trained reward model can be used in many different ways to improve the level of engagement of the finetuned chatbot model. One approach would be to use Proximal Policy Optimization (PPO) with reinforcement learning~\cite{DBLP:journals/corr/SchulmanWDRK17}, to directly update the chatbot model weights to generate responses that maximise the reward. In this work we take the first steps towards PPO by establishing a preliminary understanding of the efficacy of our reward model, but then aim to exploit it in a convenient and quick manner. Therefore, instead of updating chatbot model parameters, we apply best of $N$ sample rejection. At inference time, $N$ responses are statistically drawn from the finetuned chatbot model
\begin{equation}
    S = \{r^{1}_i, r^{2}_i, \dots, r^{N}_i \},
\end{equation}
and the response with the highest reward score is then selected and given to the user
\begin{equation}
    r_i = \underset{r \in S}{\text{argmax}} \; R(r).
\end{equation}

\section{Experiments}
\noindent\textbf{Chatbot}
The chatbot we use is the GPT-J 6B \citep{gpt-j} model fine-tuned on novels and literature\footnote{\url{https://huggingface.co/hakurei/lit-6B}}. Though dialogue data could have been used at this stage, we found that training on standard conversational corpora led to bland and uninteresting responses, while fine-tuning on literature and novels (that include screen play and dialogue) led to more captivating responses. 

\noindent\textbf{Reward Model}
Our reward models are based on either the pre-trained GPT-2 small (124M parameters), medium (355M), large (774M) and extra-large (1.5B) models. The vector representation is taken from the final position and fed through a simple classification head to get the output logits. All model weights re learned during training, where we select the model weights for the epoch which minimises the validation loss. The input to the reward model is the previous 


\noindent\textbf{Data set and Evaluation}
We use the \textit{Chai user response} dataset, a collection of 50 million partial conversations ending with a response from the chatbot, with annotations including turn number, the number of subsequent user messages, whether regeneration was requested, and user star ratings (when applicable). Each full conversation is thus represented multiple times in this dataset, once for each turn taken by the chatbot. This dataset is used to train the reward model and we publicly release this dataset to enable further research into building engaging chatbots. 

To evaluate the effectiveness of the reward models, we run live A/B experiments on the Chai Research platform. We deploy a baseline chatbot system (without the reward system) as well as the same chatbot system with an additional reward model module. We then create mutually exclusive cohorts of users (randomly drawn from both new and existing users) and assign them either to the baseline system or with the system which uses selection via reward model scores. We track the user interactions over the next twelve hours (though extend this time if there are fewer than 20000 sessions in the cohort) and track the mean conversation length. We note that since A/B tests were run at different times in the day at various points in the year, with changing user demographics and population statistics, it may not be a fair to compare the absolute performance of the A/B tests across different time periods. We therefore only consider \textbf{relative performance increase} of the system over the baseline of no reward model.

For systems where the improvement in the mean conversation length is better than the previous best reward model, we run a second more detailed retention evaluation where new users are split into two groups, each of at least 10000 users. The daily user usage is log in each day for the next thirty days, and we measure the retention statistics. Again, since external factors such as growth of the platform and changes in the user interface and may influence user usage, we only consider the relative increase in retention. 

\section{Results}
\subsection{Last response labels}
We first investigate whether a reward model can be used to improve the quality of chatbot responses, and to identify the impact of training data size as well as the initial PrLM architecture chosen. Our experiments use reward models trained with \textit{continuation pseudo labels} (K=1 for psuedo labels, N=4 during reward selection). Table \ref{tab:base_PrLm} compares the performance of RoBERTa-based reward model to a GPT-2 based reward model, with GPT-2 clearly leading to the best reward model. \textbf{can add a line here discussing hypothesis why (same base architecture) and conclusion of the experiment}. GPT-2 based reward models are therefore only consider for the rest of the paper.

\begin{table}[h]
  \small
  \centering
  \begin{tabular}{lccc}
    \toprule
    system  & \# param & train size & MCL change (\%)  \\
    \midrule
    RoBERTa dist.  & 82M  & 165k & $+2.23 \pm 1.00$ \\
    RoBERTa base   & 125M & 165k & $+0.84 \pm 1.02$ \\
    RoBERTa large  & 355M & 90k  & $+3.34 \pm 1.00$ \\
    GPT2 small     & 124M & 125k & $+8.82 \pm 1.45$ \\
    \bottomrule
  \end{tabular}
  \caption{Percentage improvement of MCL relative to not having a reward model when the GPT-2 large reward model predicts whether the conversation continues or whether the user retries the message.}
  \label{tab:base_PrLm}
\end{table}

\begin{figure}[h]
  \centering
  \includegraphics[width=\columnwidth]{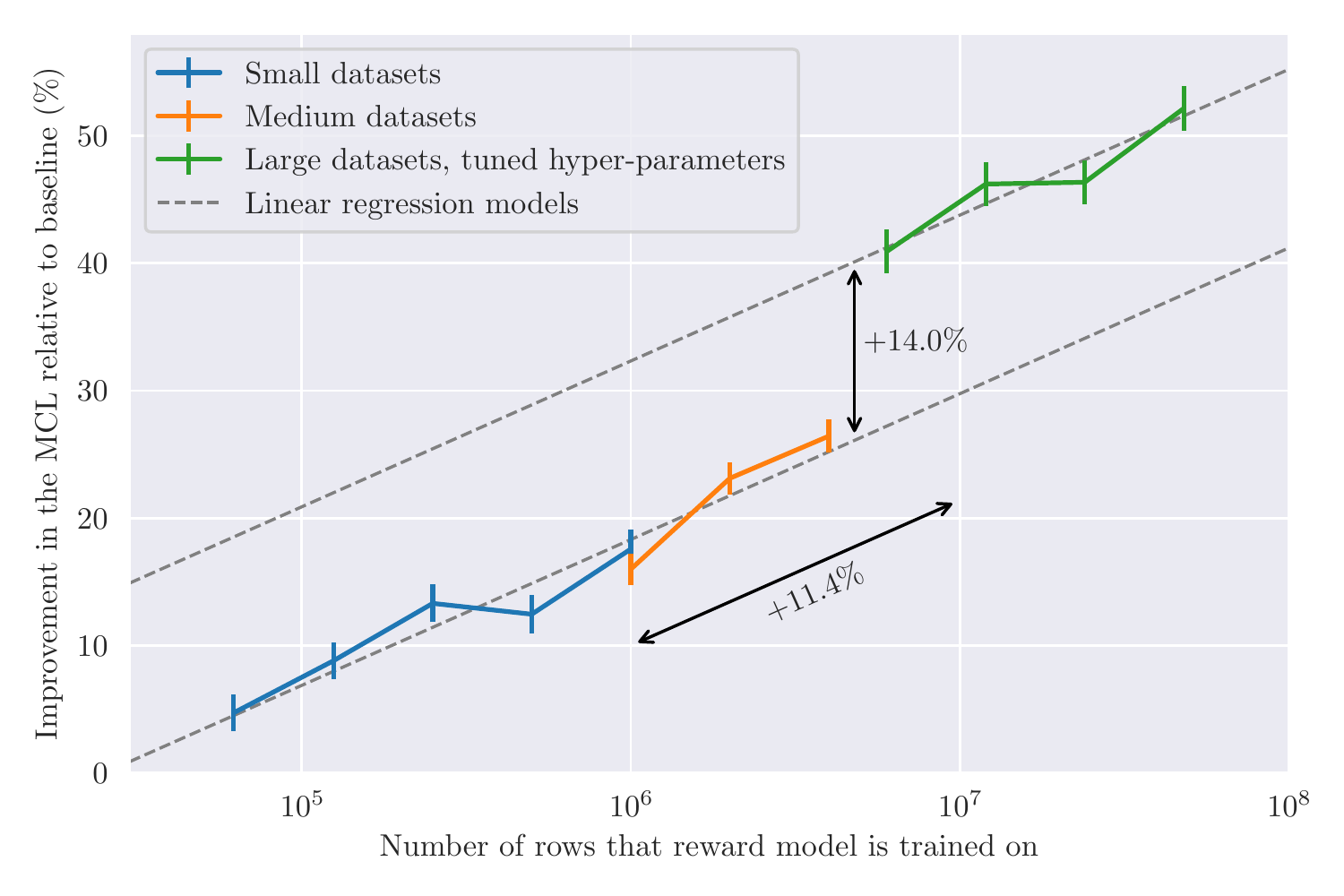}
  \caption{Scaling of the percentage improvement in Mean Conversation Length (MCL) with the dataset size that our GPT-2 reward model is trained on. Each of the three sets of datapoints corresponds to a separate A/B experiment run on the Chai platform. The grey dashed line is a linear regression fit of all three experiments, with an additional parameter describing the improvement due to tuning the hyper-parameters of the reward model context and label.}
  \label{fig:scaling_with_data}
\end{figure}

Further, scaling the training data size of the reward model was found to lead to large improvements in response quality. Figure \ref{fig:scaling_with_data} shows that by training a GPT-2 small reward model on the Chai user response data with between 62K rows to 24M rows (where a \textit{row} is a chatbot response with conversational history), we observe  a clear linear relationship between the log number of rows and MCL evaluation performance ($m=11.4$ and $c=-50.3$ found using Levenberg-Marquardt \citet{more2006levenberg}), and a remarkable 50\% increase in MCL when using the full 24M rows. This provides strong promise for using human feedback to create more engaging chatbots, where performance has not yet saturated implying that further improvements can be expected by scaling up data. Note that the 14\% increase jump in performance at 8M rows was due to a change in input formatting, where the reward model input was increased from the last three user-chatbot turns to the last 256 tokens of the conversation. Although using larger datasets clearly leads to notably better performance, in further results we limit the dataset to X M samples.

\label{sec:continuation}

\subsection{Impact of hyper-parameter choices}
We further look into the influence of the parameters such as context window size, the cut off point K, and the number of generated samples N. 

\begin{figure}[h]
  \centering
  \includegraphics[width=\columnwidth]{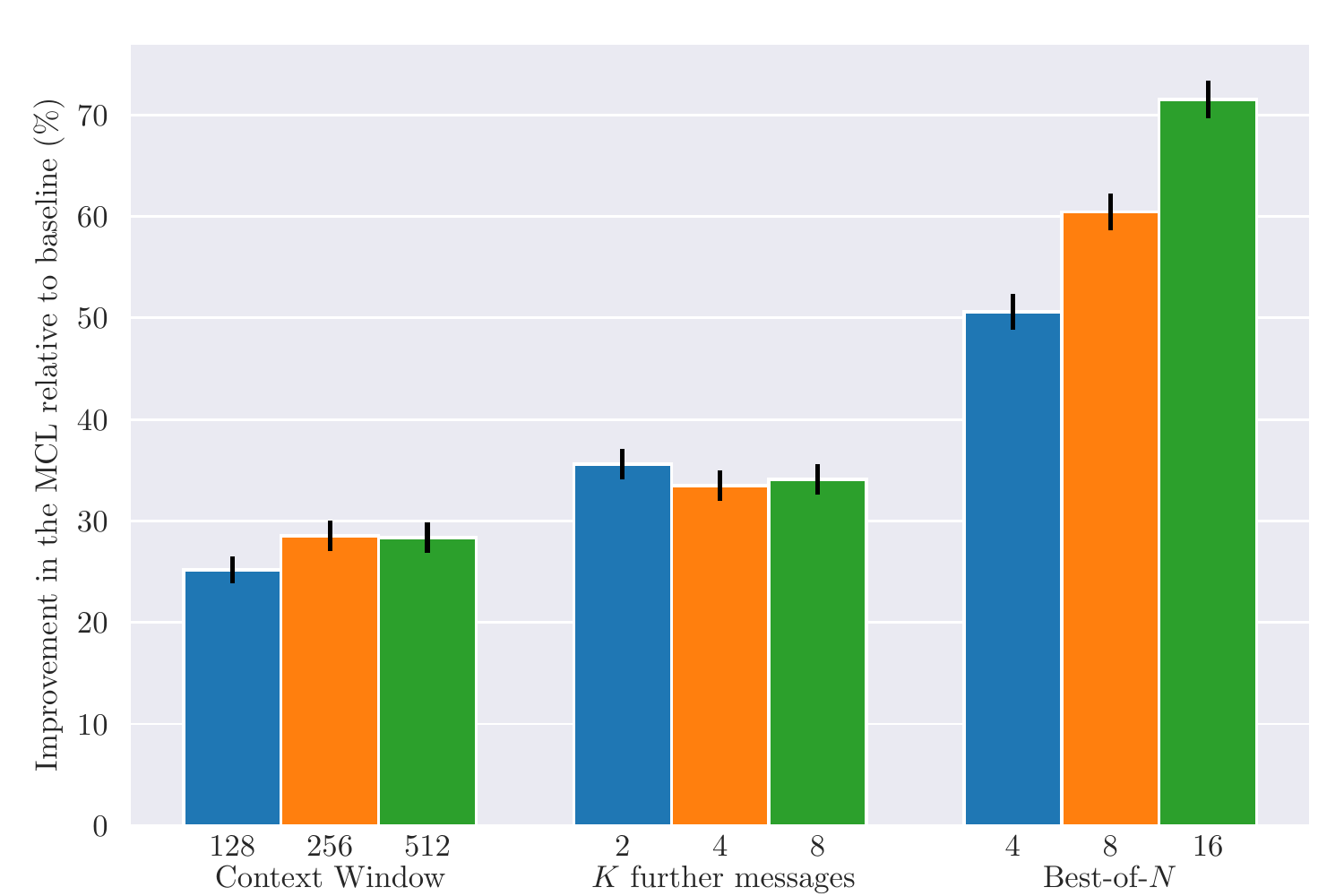}
  \caption{We ran a suite of three A/B experiments on the Chai platform exploring how the percent improvement in MCL relative to not using a reward model varied. \textbf{Left:} Increasing the number of tokens given to the reward model as context. \textbf{Middle:} Increasing the threshold number of user messages sent after the chatbot message for the conversation to count as having continued. \textbf{Right:} Increasing the number of potential responses from the chatbot that the reward model can choose from.}
  \label{fig:hyperparameter_tuning}
\end{figure}

\subsubsection*{Input context window}
Previous experiments had a reward model where the input was the final three user messages and chatbot response pairs. We investigate whether providing more information to the reward model may improve its ability in scoring responses, and consider filling up the input with final 128, 256 or 512 tokens. With $N=4$ and $K=2$, we find that using 256 tokens instead of 128 tokens increases MCL by $+2.65 \pm 1.30 \%$, but that further increasing the context length to 512 tokens decreases MCL by $-0.11 \pm 1.28 \%$.

\subsection*{Engagement Cut off Point}
In the previous section, for the \textbf{last response} pseudo label it was assumed that only the last response was of poor quality and caused the conversation to end. Therefore, only the final response was given a pseudo label of 0 (K=1). However one may alternatively believe that the final 2 responses were non-engaging (K=2), or that the final 4 were (K=4), as those responses led the user to eventually disengage. To investigate the influence of K on performance, the reward model was trained with pseudo labels using $K\in\{1, 2, 4, 8\}$ (with N=4 and a context window of 256), and it was found that $K=2$ led to a further X\% improvement in MCL, but that increasing K further was unhelpful.

\subsubsection{Number of Samples}
The reward model is used by over sampling stochastic responses and then selecting the response with the highest reward model score. It is expected that by sampling more responses than the initial decision of $(N=4)$, the quality of the returned response will be higher. This was confirmed experimentally where we found a significant increase of $+6.90 \pm 1.32 \%$ in MCL for N=8, or an increase of $+13.62 \pm X\%$ by increasing N from 4 to 16. However, doubling the number of samples also doubles the inference cost of the conversation. 

It was observed, though, that other factors such as latency can largely influence user experience and observed engagement. We found that synthetically adding one second of latency led to a decrease in MCL by $-3.01\%$, while two seconds of latency decreased by MCL by $-6.10\%$. This highlights that other factors beyond response quality can impact how engaging the user finds the system, and as a compromise between conversation quality, inference cost and latency times, we stick with the default $N=4$ in subsequent experiments. 


\subsection{Retry labels}
Previous experiments used final response pseudo labels to train reward models. However section X proposed other labels that could be used as a proxy for response quality, and this section focuses on the alternate retry labels.  We trained several reward models using retry-based pseudo labels using a range of dataset sizes (1.5M, 3M, 6M and 12M). We ran our standard A/B evaluation and found that though the retry labels would improve

\begin{table}[h]
  \small
  \centering
  \begin{tabular}{ccc}
    \hline
    Reward target  & \# of rows & MCL improvement (\%)  \\
    \hline
    Last response & 12M & $+50.87 \pm 1.65 \%$ \\
    \hline
    Retries & 12M &  $+20.59 \pm 1.48 \%$ \\
    Retries & 6M & $+15.98 \pm 1.45 \%$ \\
    Retries & 3M & $+9.00 \pm 1.40 \%$ \\
    Retries & 1.5M & $+8.58 \pm 1.41 \%$ \\
    \hline
  \end{tabular}
  \caption{Percentage improvement of MCL relative to not having a reward model when the GPT-2 large reward model predicts whether the conversation continues or whether the user retries the message.}
  \label{tab:retry_scaling}
\end{table}



\begin{table}[h]
  \small
  \centering
  \begin{tabular}{ccc}
    \hline
    Reward target  & \# of rows & MCL improvement (\%)  \\
    \hline
    Last response & 12M & $+50.87 \pm 1.65 \%$ \\
    \hline
    Retries & 12M &  $+20.59 \pm 1.48 \%$ \\
    Retries & 6M & $+15.98 \pm 1.45 \%$ \\
    Retries & 3M & $+9.00 \pm 1.40 \%$ \\
    Retries & 1.5M & $+8.58 \pm 1.41 \%$ \\
    \hline
  \end{tabular}
  \caption{Percentage improvement of MCL relative to not having a reward model when the GPT-2 large reward model predicts whether the conversation continues or whether the user retries the message.}
  \label{tab:retry_scaling}
\end{table}

\subsection{Predicting whether the conversation continues without retrying the response}
\label{sec:both}
In the previous experiment we found that having the reward model predict whether the user retries the message causes a much smaller improvement in the MCL versus predicting whether the conversation continues. This is not a surprising result, given that choosing a message that is more likely to cause the conversation to continue is more closely aligned with longer conversations than choosing a message which the user is less likely to retry. We ran a suite of experiments to explore whether predicting if the user retries leads to a reward model that improves our retention in a direction that is orthogonal to increasing MCL.

The first reward model predicts whether the conversation continues, the second predicts whether the user does not retry the message, and third predicts whether the intersection of those events occurs. In all three cases we trained a GPT-2 small reward model. We ran an A/B test of these three reward models plus the baseline of no reward model over a period of thirty days. Each was assigned a distinct cohort of new users for the duration of the experiment. We recorded the mean conversation length and retry rate over that entire period, as well as the retention of the initial cohort at days $D\in\{1,2,3,4,5,6,7,10,15,20,25,30\}$.

\begin{figure}[h]
  \centering
  \includegraphics[width=\columnwidth]{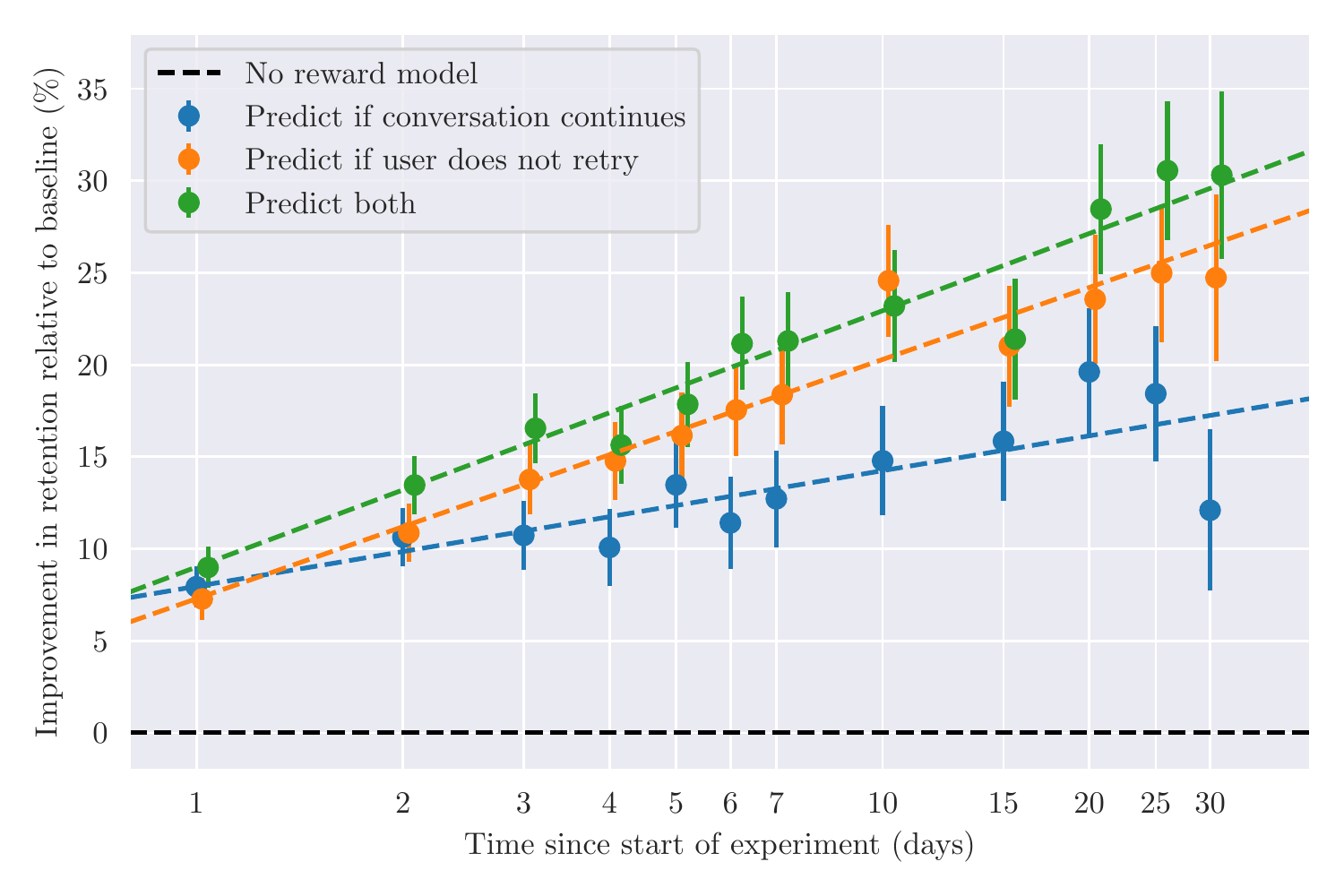}
  \caption{Percent improvement in user retention across time relative to not using a reward model. We show three reward models based on GPT-2 small that were each trained to predict a different event: i) whether the conversation continues, ii) whether the user does not retry the last chatbot message, iii) whether both of those events occur. The dashed lines are linear regression fits to the corresponding data-points, showing the log-linear relationship between improvements to our reward models and improvements in retention.}
  \label{fig:retention}
\end{figure}

In Fig. \ref{fig:retention} we show the percentage improvement in retention relative to not using a reward model over time. We find that in all three cases the percentage improvement is consistent with increasing linearly with the log of the number of days since the start of the experiment. This suggests that the fraction of users retained decays exponentially with time, and that using a reward model decreases the rate of the decay. By day 30, we find that predicting whether the conversation continues increases retention by $+12.1 \pm 4.4 \%$, predicting whether the user does not retry increases retention by $+24.7 \pm 4.5 \%$, and predicting whether both events occur increases retention by $+30.3 \pm 4.5 \%$.

In Tab. \ref{tab:comparing_metrics} we show the percent improvement in the MCL, retry rate and retention of rewards models targeting conversation continuation, the user not retrying the message and the intersection of both, relative to the baseline of not using a reward model.

\begin{table}
  \centering
  \begin{tabular}{cccc}
    \hline
    & \multicolumn{3}{c}{Improvement (\%)} \\
    Reward target & MCL & Retry & Retention\\
    \hline
    Continues & $+23.6$ & $-8.7$ & $+12.1$  \\
    Retries & $+3.6$ & $-7.9$ & $+24.7$  \\
    Both & $+16.8$ & $-8.2$ & $+30.3$  \\
    \hline
  \end{tabular}
  \caption{Improvement in metrics observed in the A/B experiment comparing different reward model target labels.}
  \label{tab:comparing_metrics}
\end{table}

\subsection{Predicting the user rating}
Another possible label to target with the reward model is whether the message received a user rating of $s$ stars or greater. Tab. \ref{tab:stars} shows the distribution of star ratings by Chai users. We trained three reward models to predict whether a chatbot response in the context of the previous conversation would receive a rating of two or more stars, three or more stars, or four stars. An A/B experiment comparing each of these to the baseline of no reward model found that the two stars or more reward model improved the MCL by $+8.70 \pm 2.54 \%$, the three stars or more reward model improved the MCL by $+9.81 \pm 2.57 \%$, while the four stars reward model improved the MCL by only $+1.24 \pm 2.47 \%$. These results suggest that it is more important for the reward model to avoid low-scoring one star or two star messages than for it to prioritise four star messages. These MCL improvements are much lower than the improvements from using either conversation continuation or the continutation and no retry label from the previous section, however exploring an ensemble of all three labels is a promising future research direction.

\begin{table}
  \centering
  \begin{tabular}{crr}
    \hline
    User rating    & Fraction  & Survival fraction\\
    \hline
    \faStar\faStarO\faStarO\faStarO & 1.6\% & 100\% \\
    \faStar\faStar\faStarO\faStarO & 3.1\% & 98.4\% \\
    \faStar\faStar\faStar\faStarO & 10.8\% & 95.1\% \\
    \faStar\faStar\faStar\faStar & 84.3\% &  84.3\% \\
    \hline
  \end{tabular}
  \caption{The distribution of user ratings of chatbot messages on the Chai platform is top-heavy, with very few messages receiving one star ratings.}
  \label{tab:stars}
\end{table}

\subsection{Reward model generalisation}

In the previous sections we explored using a GPT-2 small (124M paramaters) reward model to rank the responses of a fine-tuned GPT-J (6B parameters) chatbot. In this section we will explore whether the improvement in MCL generalises to both i) larger GPT-2 reward models and ii) a differently fine-tuned GPT-J (6B parameters, Pygmalion\footnote{\url{https://huggingface.co/PygmalionAI/pygmalion-6b}}) chatbot.

We fine-tuned the four GPT-2 models (small, 124M parameters; medium, 355M; large, 774M; extra-large, 1.5B) to predict the `both' label from Sec. \ref{sec:both} (does the conversation continue for another two user messages and does the user not retry this chatbot response?) on 12M rows with a 256 token context window. We ran an A/B experiment comparing these four models to the baseline without a reward model. We show the percent improvement in the MCL in Tab. \ref{tab:model_size}. Fitting a linear regression between $\log_{10}(\#\;\mathrm{of}\;\mathrm{parameters})$ and the MCL improvement, we find that increasing the number of parameters by a factor of ten gives an MCL improvement of $+5.0 \pm 1.1 \%$. We can compare this to the $+11.4 \pm 1.0 \%$ increase in MCL improvement from increasing the dataset size that the reward model is trained on by a factor of ten that we found in Sec. \ref{sec:continuation}. We conclude that increasing the reward model fine-tuning dataset set size by a factor of ten is more than twice as effective at increasing the MCL improvement than increasing the model size by a factor of ten.

\begin{table}
  \centering
  \begin{tabular}{ccc}
    \hline
    \# of parameters & MCL improvement (\%)\\
    \hline
    124M & $+27.95 \pm 1.29 \%$  \\
    355M & $+29.68 \pm 1.30 \%$  \\
    774M & $+32.95 \pm 1.32 \%$  \\
    1.5B & $+32.82 \pm 1.31 \%$  \\
    \hline
  \end{tabular}
  \caption{Percent improvement in mean conversation length from increasing the size of the GPT-2 reward model.}
  \label{tab:model_size}
\end{table}

We ran a further A/B experiment exploring the performance of our own fine-tuned GPT-J chatbot versus the Pygmalion fine-tuned GPT-J chatbot, with and without the GPT-2 small reward model from the previous experiment. We found that Pygmalion GPT-J chatbot gave an MCL improvement over the fine-tuned GPT-J model used previously in this paper of $+16.40 \pm 2.71 \%$, without using a reward model. Using the the GPT-2 small reward model improved the performance of our fine-tuned GPT-J by $+36.87 \pm 2.89 \%$, while the improvement of the Pygmalion GPT-J was increased to $+54.33 \pm 3.08 \%$. We modelled the percent improvement to the MCL in these three cases as
\begin{equation}
    y = b \mathbbm{1}_{\mathrm{Pygmalion}} + c\mathbbm{1}_{\mathrm{reward}\;\mathrm{model}}, \label{eq:pygmalion}
\end{equation}
where $b$ and $c$ model the separate additive contribution to the improvement from using Pygmalion versus using our GPT-2 small reward model. We find the values of these parameters to be $b = 16.71 \pm 0.48$ and $c = 37.22 \pm 0.50$. In other words, using the Pygmalion chatbot increases the MCL by $+16.7\%$, while using our reward model increases the MCL by $+37.2\%$. We can therefore conclude that the reward model improvement has generalised between the two differently fine-tuned GPT-J chatbots.
\section{Conclusion}
In this work, we focus on developing chatbots that are highly engaging and entertaining for users. We demonstrate through extensive A/B tests across large user groups that training reward models with human feedback followed by response selection, leads to chatbots with longer average user interactions and higher user retention. We propose intuitive evaluation metrics, namely the mean conversation length and user retention, and investigate a range of pseudo labels that can be used to identify captivating responses that can be used to train a reward model that can score generated responses. Using our best reward model for response selection, we show that the user retention of a GPT-J 6B language model increases by over 30\%, highlighting the effectiveness of using natural human-chatbot interactions for developing highly engaging chatbots.






\bibliography{anthology,custom}
\bibliographystyle{acl_natbib}

\appendix
\newpage
\section{Appendices}

\subsection{Conversation lengths}
\label{sec:conversation_length}

\begin{figure}
  \centering
  \includegraphics[width=\columnwidth]{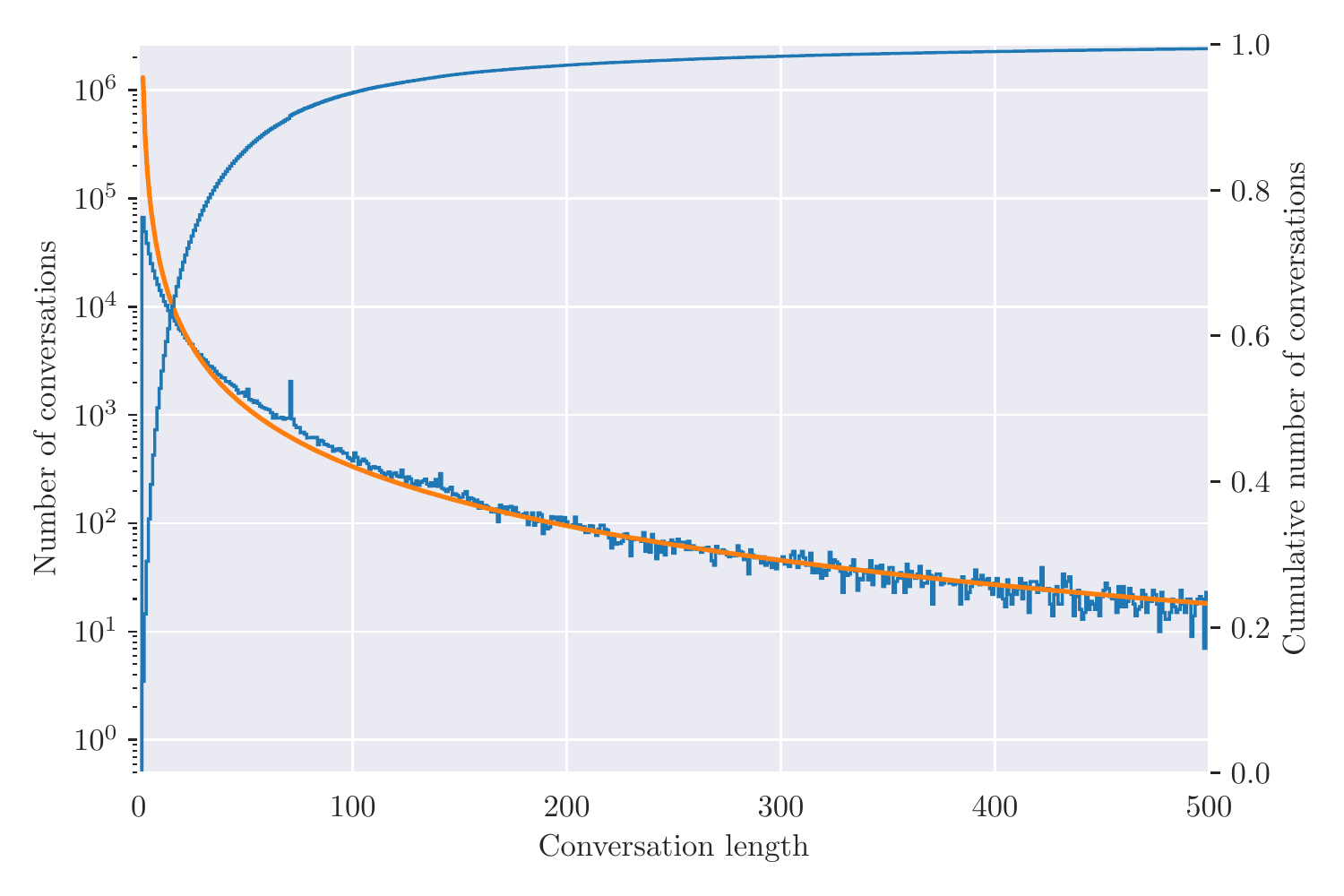}
  \caption{Distribution (left) and cumulative distribution (right) of the lengths of conversations on the Chai platform on the 25\textsuperscript{th} October 2022. The orange line shows a Zeta distribution with arbitrary scaling and a power-law slope of -1.8 to illustrate the long-tail of conversation lengths.}
  \label{fig:conversation_length}
\end{figure}

There is a wide range in the lengths of the conversations on the Chai Platform. In Fig. \ref{fig:conversation_length} we show the distribution of conversation lengths of the 531,044 conversations on the 25\textsuperscript{th} October 2022. We only record conversations that consist of at least two messages: the initial message from the chatbot and a response from the user. The distribution follows a decaying power-law with 50\% of conversations lasting for ten or fewer messages, while 1.2\% of conversations last for more than 500 messages and 0.1\% of conversations last for more than 5,000 messages. The spike in the distribution at conversation lengths of 70 is due to users that are not paying for the premium service being restricted to at most 70 messages per day, and those users using all of their messages in a single conversation.

In Fig. \ref{fig:conversation_length} we also show an illustrative power-law with a slope of -1.8 which well-describes the tail of long conversations. The Zeta distribution (a power law distribution with negative exponent on the positive-valued integers) has an undefined variance when the exponent is greater than or equal to -3. This has a practical consequence for measurements of the mean conversation length: the sample mean of values drawn from a distribution with undefined variance does not converge to the population mean. For the remainder of this work we re-define the mean conversation length to refer to the mean of only those conversations that were at most 100 messages long.

\end{document}